\documentclass[accepted]{uai2024}
\usepackage[american]{babel}
\usepackage{natbib} % has a nice set of citation styles and commands
\usepackage{mathtools} % amsmath with fixes and additions
\usepackage{booktabs} % commands to create good-looking tables
\usepackage{tikz} % nice language for creating drawings and diagrams
\newtheorem{theorem}{Theorem}[section]

\newtheorem{lemma}[theorem]{Lemma}
\newtheorem{corollary}[theorem]{Corollary}

\newtheorem{assumption}[theorem]{Assumption}

\usepackage[textsize=tiny]{todonotes}

\usepackage{amssymb}
\usepackage[formats]{listings}
\title{Normalizing Flows for Conformal Regression}
\author[1]{Nicolo~Colombo}
\affil[1]{%
    {\tt nicolo.colombo@rhul.ac.uk}\\
    
    Computer Science Department\\
    Royal Holloway, University of London\\
    Egham, Surrey, UK
}

\begin{document}
\maketitle

\begin{abstract}
Conformal Prediction (CP) algorithms estimate the uncertainty of a prediction model by calibrating its outputs on labeled data.  The same calibration scheme usually applies to any model and data without modifications. The obtained prediction intervals are valid by construction but could be inefficient, i.e. unnecessarily big, if the prediction errors are not uniformly distributed over the input space.

We present a general scheme to localize the intervals by training the calibration process. The standard prediction error is replaced by an optimized distance metric that depends explicitly on the object attributes.  Learning the optimal metric is equivalent to training a Normalizing Flow that acts on the joint distribution of the errors and the inputs.  Unlike the Error Reweighting CP algorithm of \cite{papadopoulos2008normalized}, the framework allows estimating the gap between nominal and empirical conditional validity. The approach is compatible with existing locally-adaptive CP strategies based on reweighting the calibration samples and applies to any point-prediction model without retraining.

\end{abstract}
\section{Introduction}
\label{section introduction}
In natural sciences, calibration often refers to comparing measurements of the same quantity made by a new device and a reference instrument.\footnote{The International Bureau of Weights and Measurements defines calibration as the
\emph{"operation that, under specified conditions, in a first step, establishes a relation between the quantity values with measurement uncertainties provided by measurement standards and corresponding indications with associated measurement uncertainties (of the calibrated instrument or secondary standard) and, in a second step, uses this information to establish a relation for obtaining a measurement result from an indication."}
}  
In data science, calibrating a model means quantifying the uncertainty of its predictions.
Parametric and non-parametric methods for model calibration have been proposed in the past.
Examples of trainable post hoc approaches are Platt scaling \citep{platt1999probabilistic}, Isotonic regression \citep{zadrozny2002transforming}, and Bayesian Binning \citep{naeini2015obtaining}. 
Here we focus on regression problems, where data objects have an attribute, $X \in {\cal X}$, and a real-valued label, $Y \in {\mathbb R}$.
The model is a \emph{point-like} predictor of the most likely label given its attribute, i.e. $f(X) \approx {\rm E}(Y|X)$.
Calibrating $f$ would promote $f(X)$ to a Prediction Interval (PI), i.e. a \emph{subset of the label space}, $C \subseteq {\mathbb R}$, that contains the unknown label, $Y$, with lower-bounded probability.
Given a target \emph{confidence level}, $1 - \alpha \in (0, 1)$, $C$ is \emph{valid} if it contains the unknown label with probability at least $1 - \alpha$, i.e. if ${\rm Prob}(Y \in C) \geq 1-\alpha$. 

Conformal Prediction (CP) is a frequentist approach for producing valid PIs without making assumptions on the data-generating distribution, $P_{XY}$, or the prediction model, $f$ \citep{vovk2005algorithmic, shafer2008tutorial}.
PIs are obtained by evaluating the \emph{conformity} between the predictions and the labels of a \emph{calibration set}. 
The evaluation is based on a \emph{conformity function}, e.g. the absolute residual, $a(Y, f(X)) = |Y - f(X)|$.
Validity is guaranteed automatically by the properties of finite-sample empirical distributions. 
Different conformity functions, however, may produce non-equivalent PIs.
Several criteria have been proposed to assess their \emph{efficiency} \citep{vovk2016criteria}.
For real-valued labels, a straightforward criterion is the average size, ${\rm E}(|C|)$.
If the model performs uniformly over the support of $P_{XY}$, the PIs obtained using $a = |Y - f(X)|$ have minimal average size.
If the data are heteroscedastic, input-adaptive techniques may increase the PI efficiency because their size changes according to the performance of $f$, e.g. the prediction band shrinks where $|Y - f(X)|$ is small and grows where $|Y-f(X)|$ is large.

\subsection{Outline}
\label{section outline}
We obtain \emph{input-adaptive} PIs by learning calibration functions that depend on the object attributes explicitly.
For simplicity, we assume the calibration samples, $(X_1, Y_1), \dots, (X_N, Y_N)$, and the test object, $(X_{N+1}, Y_{N+1})$, are independently drawn from the same \emph{joint distribution}, i.e. $(X_n, Y_n) \sim P_{XY}$.
The method applies with minor changes if the samples are only \emph{exchangeable} \citep{vovk2005algorithmic}. 
Given a conformity function, $a:{\mathbb R}^2 \to {\mathbb R}$, and a confidence level, $1-\alpha \in (0, 1)$, CP consists of two main steps, 
\begin{enumerate}
    \item computing the $(1-\alpha)$-th \emph{sample quantile}, $Q_A$, of the calibration scores, $A_n=a(Y_n, f(X_n))$, $n=1, \dots, N$, and 
    \item accept all possible test labels, $y \in {\mathbb R}$, for which $a(y, f(X_{N+1}))$ is smaller than $Q_A$.
\end{enumerate}
If $A_n = |Y_n-f(X_n)|$, the PI at $X_{N+1}$ is the interval $C_A = [f(X_{N+1}) - Q_A, f(X_{N+1}) + Q_A] \subseteq {\mathbb R}$.
Since $Q_A$ is the $(1-\alpha)$-th sample quantile of $\{ A_n\}_{n=1}^N$ and  calibration and test samples are i.i.d., $C_A = \{ y \in {\mathbb R}, a(y, f(X_{N+1}))\leq Q_A\}$ guarantees ${\rm Prob}(Y_{N+1} \in C_A) \geq 1 - \alpha$.
We say that $C_A$ is \emph{marginally valid} because $Q_A$ approximates the quantile of the marginal distribution $P_A = \sum_{XY}P_{AXY}$\footnote{Technically, marginal validity depends on the joint distribution of the calibration and test samples, i.e. ${\rm Prob}(Y_{N+1} \in C_A) = P_{X_{N+1}Y_{N+1}X_1Y_1\dots X_NY_N}(Y_{N+1} \in C_A)$.}.
In particular, there is no conditioning on the test input, $X_{N+1}$ \citep{vovk2012conditional}.
PIs with input-conditional coverage, ${\rm Prob}(Y_{N+1} \in C_{A}|X_{N+1})$ \emph{cannot} be obtained with finite data and without certain regularity assumptions on the data distribution \citep{lei2012distribution, vovk2012conditional, foygel2021limits}.
Approximating distribution-free conditionally-valid PIs is the goal of an active research stream (see Section \ref{section related work}). 
Existing methods are mostly based on importance-sampling techniques that temporarily break the data exchangeability \citep{lin2021locally, tibshirani2019conformal, guan2023localized}\footnote{In \cite{lin2021locally, tibshirani2019conformal, guan2023localized}, the sample quantile of $\{ A_n\}_{n=1}^N$ is replaced by the quantile of an importance-sampling estimate of the empirical input-conditional distribution, $P_{A|X} \approx \sum_{n=1}^N w_n(X) {\bf 1}(A =A_n)$, where $\sum_{n=1}^N w_n(X)  = 1$ and $w_n(X)$ depends on $X$ through a predefined function.}.
Our strategy is to preserve exchangeability at all times but change the definition of the conformity function, i.e. to replace $a$ with $b = b(a(Y, f(X)), X) \in {\cal B}$ and apply $b$ unconditionally to the calibration and test samples.
Data exchangeability holds automatically provided $b$ is trained on a separate set.
As in standard CP, we use the \emph{transformed calibration samples}, $B_n = b(a(Y_n, f(X_n)), X_n)$, $n=1, \dots N$, to compute a \emph{$B$-space threshold}, $Q_B$, and build PIs that are marginally valid, i.e. of \emph{constant} size, over ${\cal B}$.
Local adaptability arises when the PIs, $C_B = \{ y \in {\mathbb R}, b(a(y, f(X_{N+1})), X_{N+1}) \leq Q_B \}$, are \emph{mapped back} to the label space (by inverting $b$).

This work addresses the following problem, 
\begin{center}
    \emph{What transformations of the conformity function improve CP adaptivity? How can we optimize a transformation using a separate training set (from the same task)?}
\end{center}
We start by interpreting $b$ as a \emph{Normalizing Flow} (NF), i.e. a coordinate transformation that maps a \emph{source} distribution, $P$, into a \emph{target} distribution, $P'$ \citep{papamakarios2021normalizing}.
In our case, the source distribution is the joint distribution of the conformity scores and the object attributes, $P_{AX}$.
The target is a factorized distribution, $P_{BX} = U_{B} P_X$, where $U_{B}$ is an arbitrary univariate distribution.
In the $B$-space, the PIs are marginally valid and have constant size by construction. 
Maximal efficiency is guaranteed because the joint distribution factorizes, which implies $P_{B|X} = U_B$ for all $X$ and the equivalence between marginally and conditionally valid PIs.
The practical problem is to enforce the factorization given the available data. 
The idea is to train $b$ by maximizing the likelihood of the transformed samples under $U_{B}$. 
When $b$ is invertible (in its first argument and for any $X_{N+1}$), the PIs are $C_B = \{ y \in {\mathbb R}, a(y, f(X_{N+1})) \leq \xi_{X} \}$, $\xi_{X} = b^{-1}(Q_B, X_{N+1})$, with $b^{-1}$ defined by $b^{-1}(b(A, X), X) = A$.
Intuitively, this produces locally adaptive PIs because $\xi_{X}$ approximates the unavailable conditional quantile $Q_{A|X}$.
The approximation error depends on the distribution distance between the source and the target distributions, $P_{b(A, X), X}$ and $U_{B} P_X$. 

\subsection{An example}
Let $P_X={\rm Uniform}({\cal X})$ be the uniform distribution over ${\cal X}=[0, 1]$ and $(X_{1}, Y_{1}), \dots, (X_{N}, Y_{N}), (X_{N+1}, Y_{N+1}) \in {\cal X} \times {\mathbb R}$ a collection of i.i.d. random variables from
\begin{align}
\label{toy model}
    &P_{XY}=P_{Y|X}P_X, \\
    &P_{Y|X} \sim  \left({\bf 1}_{<0.5} + \xi \ {\bf 1}_{>0.5} \right){\cal N}(0,1) \nonumber
\end{align}
where ${\bf 1}_{<\frac12} = {\bf 1}(X < 0.5)$, ${\bf 1}_{>0.5} = {\bf 1}(X > 0.5)$, and $\xi = 5$.
Assume we have the best-possible prediction model, i.e. $f(X) = {\rm E}(Y|X) = 0$, for any $X \in {\cal X}$.
Let $a(Y, f(X)) = |Y- f(X)| = |Y|$ be the conformity measure and $A_n=|Y_n|$ the corresponding conformity scores, $n=1, \dots N+1$.
Choose a target confidence level, $1-\alpha \in (0, 1)$, and let $Q_A$ be the $(1-\alpha)$-th sample quantile of $\{A_n\}_{n=1}^N$, i.e. its $m_*$-th smallest element, $m_*=\lceil(1 - \alpha)(N+1)\rceil$.
If $N=100$ and $\alpha=0.05$, we have $m_*= 96$.
The conformity scores, $A_1, \dots, A_{N+1}$, are i.i.d. random variables because $(X_n, Y_n)$ are i.i.d. 
For any $X_{N+1}$, the marginal PI is $C_{A}=[f(X_{N+1})-Q_A, f(X_{N+1}) + Q_A] =  [-Q_A, Q_A]$, i.e. $C_{A}$ has the same width over the entire input space ${\cal X} = [0, 1]$.

\begin{figure}[t]
    \centering
    \includegraphics[scale=.5]{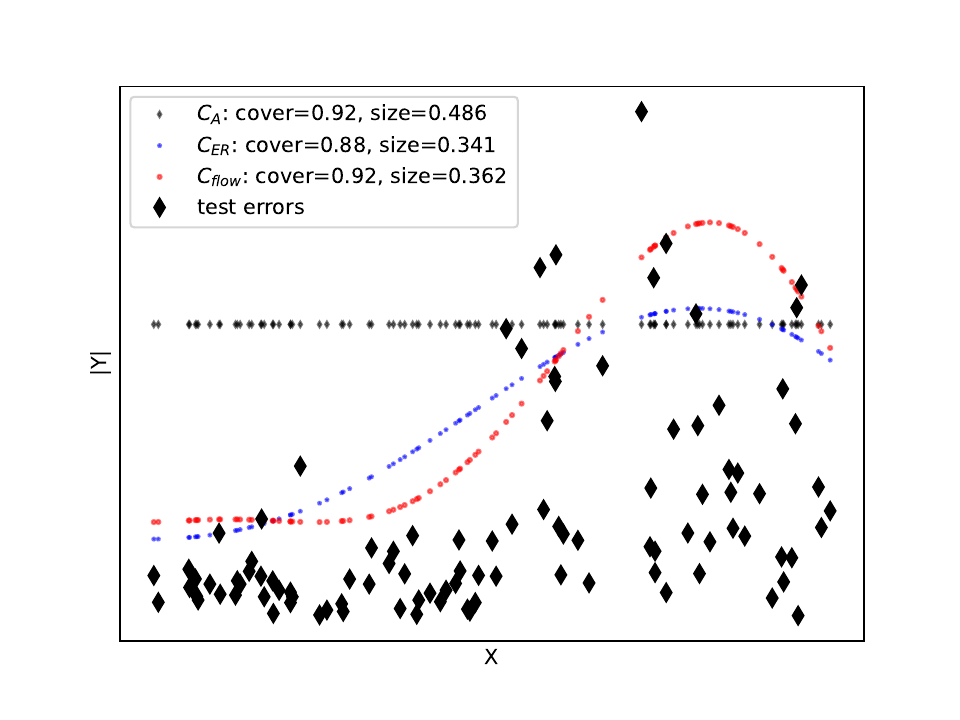}
    \caption{A test sample of conformity scores (black diamonds) and the upper bound of the marginal PIs (black dots) and the adaptive PIs obtained through the ER CP algorithm of \cite{papadopoulos2008normalized} (blue dots) and the NF approach (red dots).
    The nominal confidence level is $1-\alpha = 0.9$ for all algorithms.}
    \label{figure toy example 1}
\end{figure}

Constant uncertainty does not correspond to the true model's prediction error (see Figure \ref{figure toy example 1}).
The data are heteroscedastic because $P_{Y|X}$ in \eqref{toy model} depends on $X$ explicitly.
As the calibration samples and $(X_{N+1}, Y_{N+1})$ are all drawn from $P_{XY} = P_{Y|X} {\rm Uniform}([0, 1])$, the test score, $a(Y_{N+1}, f(X_{N+1}))$, is smaller than $A_{m*}$ with probability $\frac{m*}{N+1}$, i.e. ${\rm Prob}(Y_{N+1} \in C_{A})=\frac{m*}{N+1}$.
Figure \ref{figure toy example 1} shows that constant-size marginal PIs are valid but \emph{inefficient}.
In particular, $C_{A}$ is too large when $X_{N+1}<0.5$ and too small when $X_{N+1}>0.5$.
An adaptive CP algorithm should output PIs that are smaller or larger than $C_{A}$ when $X_{N+1}<0.5$ or $X_{N+1}>0.5$.

\begin{figure}[t]
    \centering
    \includegraphics[scale=.5]{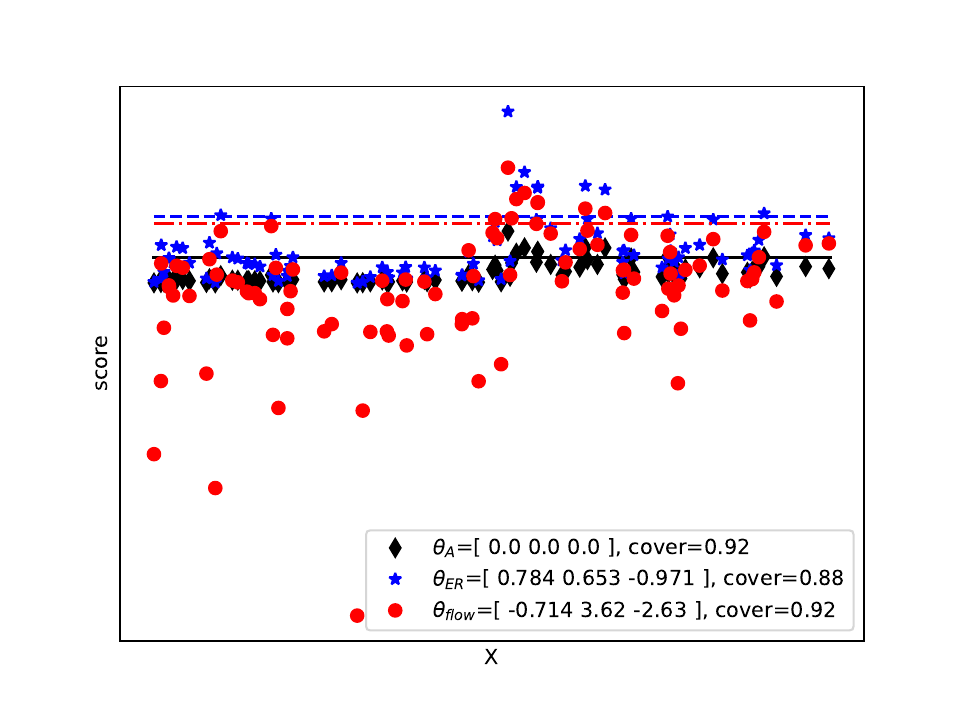}
    \caption{A calibration sample of the original conformity scores (black diamonds) and the scores obtained by transforming them with $b_{ER}$ (blue stars) and $b_{flow}$ (red dots). The solid and dashed lines represent the corresponding $(1 - \alpha)$-th sample quantiles, $1-\alpha = 0.9$.
    }
    \label{figure toy example 2}
\end{figure}

We aim to learn a \emph{locally adaptive conformity functions}, $b = b(A, X)$, that produces these adaptive PIs automatically, i.e. without partitioning the input space and using the standard CP procedure described in Section \ref{section outline}. 
Let $B_n = b(A_n, X_n) \in {\cal B}$ and $Q_B$ be the sample quantile of $\{ B_n\}_{n=1}^N$.
In ${\cal B}$, PIs are defined as in Section \ref{section outline}, i.e. $C_B = \{y \in {\mathbb R}, b(|y|, X_{N+1}) \leq Q_B\}$ and have constant size (see Figure \ref{figure toy example 2}).
Assuming $b$ is monotonic in $A_n$, there exist $b^{-1}$ such that $b^{-1}(b(A, X), X) = A$.
The inverse transformation, $b^{-1}$, can be used to map $C_B$ back to the label space, i.e. to rewrite the PIs as $ \{y \in {\mathbb R}, |y| \leq b^{-1}(Q_B, X_{N+1})\}$.
For improving PI efficiency, we need a $b$ such that $b^{-1}(Q_B, X_{N+1})$ is smaller than $Q_A$ for $X_{N+1}<0.5$ and larger than $Q_A$ for $X_{N+1}>0.5$.

Similar to the Mondrian CP algorithm \citep{vovk2005algorithmic}, we split $\{(A_n, X_n)\}_{n=1}^N$ into $D_{<0.5} = \{ (A_n, X_n), X_{n} < 0.5 \}_{n=1}^N$ and $D_{>0.5}=\{ (A_{n}, X_n), X_n>0.5 \}_{n=1}^N$.
Since $Y_n|X_n \sim {\cal N}(0, 1)$ for all $(Y_n, X_n) \in D_{< 0.5}$ and $Y_n|X_n  \sim {\cal N}(0, 5)$ for all $X_n \in D_{> 0.5}$, the quantile of $P_{A|X}$ can be written as $Q_{A|X} = {\bf 1}_{<0.5} Q_{A|X<0.5} + {\bf 1}_{>0.5} Q_{A|X>0.5}$, where $Q_{A|X<0.5}$ is the $m_*$-th smallest elements of $D_{<0.5}$, $m_{*}=\lceil(1 - \alpha)(|D_{>0.5}|+1)\rceil$ (idem for $X<0.5$).
As expected, the conditional quantile depends on $X$ through ${\bf 1}_{<0.5}={\bf 1}(X <0.5)$ and ${\bf 1}_{>0.5}={\bf 1}(X >0.5)$.
This implies the conditionally-valid PIs, $C_{A|X} = [-Q_{A|X}, Q_{A|X}]$, will depend on the location of the test object $X_{N+1}$, i.e. on whether $X_{N+1}<0.5$ or $X<0.5$. 
In this special case, the conditionally valid PIs for $X<0.5$ and $X>0.5$ are equivalent to the marginal PIs of the regions $[0,0.5]$ and $[0.5, 1]$. 
Partitioning the calibration data is optimal if i) the sample size is large enough and ii) we know the data generating distribution.
Otherwise, we need a more general approach. 

Let 
\begin{align}
\label{toy phi}
   &b_{flow} = \log\left(\frac{A}{\gamma + |g(X)|^2} \right), \quad  \\
   &g(X)= \theta_1 X + \theta_2 X^2\ + \theta_3 X^3 \nonumber 
\end{align} 
where $\theta = (\theta_1,\theta_2, \theta_3) \in {\mathbb R}^3$ is a free parameter and $\gamma = 0.01$.
For any $X$ and $\theta$, $b_{flow}(A, X)$ is a monotonic (and hence invertible) function of $A=|Y|$.
Analogously, let $b_{ER} = \frac{A}{\gamma + |g(X)|^2}$ as in the Error Reweighted (ER) CP algorithm of \cite{papadopoulos2008normalized}.
$b_{ER}$ is also a monotonic and invertible function of $A=|Y|$.
Let $\{b_{flow}(A_n,X_n)\}_{n=1}^N$  and $\{b_{ER}(A_n,X_n)\}_{n=1}^N$ be the \emph{transformed calibration sets} obtained using $b_{flow}$ and $b_{ER}$.
To compare our strategy and the ER algorithm, we look at the efficiency of $b_{flow}$ and $b_{ER}$ when $\theta_{flow}$ and $\theta_{ER}$ are trained through the proposed NF scheme or the error-fitting heuristic of \cite{papadopoulos2008normalized}. 
Figure \ref{figure toy example 2} shows a sample of the original calibration scores $\{A_n\}_{n=1}^N$ and the transformed scores obtained through $b_{flow}$ and $b_{ER}$ when $\theta_{flow} \neq \theta_{ER}$ are optimized following the corresponding strategies on a separate training data set. \footnote{As $\log(t)$ is a monotonic and input-independent transformation, the PIs obtained from $b_{ER}$ and $b_{flow} = \log \circ b_{ER}$ are equivalent if we use the same localization function, e.g. if we set $\theta = \theta_{flow} = \theta_{ER}$.}

Let $Q_{B}$ be the $(1-\alpha)$-th (marginal) sample quantile of $\{ B_{n}=b_{flow}(A_n,X_n)\}_{n=1}^N$.
By definition,  $Q_{B} = \log \left(A_{n_*}(\gamma + |g(X_{n_*})|^2)^{-1}\right)$, for some $n_*$ such that $ \lceil (1 - \alpha) (N + 1)\rceil$ elements of $\{ B_n\}_{n=1}^N$ are smaller than or equal to $Q_B$. 
In general, since $b_{flow}$ depends on $X$ explicitly, this does not imply there are $ \lceil (1 - \alpha) (N + 1)\rceil$ elements of $\{ A_n\}_{n=1}^N$ smaller than or equal to $A_{n_*}$.
The exchangeability of $B_{n}$ and $B_{N+1} = b_{flow}(|Y_{N+1}|, X_{N+1})$ guarantees the validity of the $B$-space PIs, i.e. ${\rm Prob}(B_{N+1} \leq Q_{B}) = \frac{n_*}{N+1}$.
The validity of the corresponding label-space PIs, ${\rm Prob}\left(|Y_{N+1}| \leq e^Q_{B}(\gamma + |g(X_{N+1})|^2) \right) = \frac{n_*}{N+1}$, follows from the monotonicity of $b_{flow}$.  \footnote{We use ${\rm Prob}(B_{N+1} \leq Q_{B}) = {\rm Prob}(A_{N+1} \leq b_{flow}^{-1}(Q_{B}, X_{N+1})) = {\rm Prob}(|Y_{N+1}| \leq e^Q_{B}(\gamma + |g(X_{N+1})|^2))= {\rm Prob}(Y_{N+1} \in C_B)$.
}.
Similar arguments apply to $b_{ER}$. 

The above holds for any $X_{N+1}$ and any $\theta$.
We aim to choose a $\theta$ that improves the efficiency of $C_B$, e.g. reduces its average size.
In \cite{papadopoulos2008normalized}, $\theta$ would be tuned to make $g(X)$ a model of the conditional residuals, i.e. $\theta_{ER} = \arg \min_\theta \sum_{n'=1}^N |Y_n^2 - |g(X_n)|^2|^2$, where $\{(A_{n'},X_{n'})\}_{n'=1}^N$ is a separate \emph{calibration-training set} of labeled samples. 
In this work, we interpret $(A, X) \to (b_{flow}(A, X), X)$ as an NF acting on the joint distribution of the conformity scores and the inputs, $(A, X) \sim P_{AX}$.
$b_{flow}$ is then trained by maximizing the likelihood of $(B, X)$ under a target factorized distribution, $P_{BX} = U_{B}P_X$.
If the NF transforms $(A, X)$ into $(B, X) \sim P_{BX}$ exactly, the obtained marginally-valid $B$-space PIs have maximal efficiency because $P_{B|X} = U_B = \sum_X P_{BX}$ for all $X$.
The choice of $U_B$ is arbitrary, provided its support is compatible with the transformation class, e.g. choosing $U_B = {\rm Uniform}([0, 1])$ would not be ideal in this case because  $b_{flow} \in {\mathbb R}$.
We choose $U_B={\cal N}(0, 1)$ instead and let 
\begin{align}
\label{toy density}
    &\theta_{flow} = {\rm arg} \min_\theta  \sum_{n'=1}^N |b_{flow}(A_{n'}, X_{n'})|^{2} 
\end{align}
where we use $u_B \propto \exp^{-\frac{B^2}{2}}$ and can drop the transformation Jacobian, $\partial_A b_{flow} = \frac{1}{A}$, because it does not depend on $\theta$.
As for $b_{ER}$, $\{(A_{n'},X_{n'})\}_{n'=1}^N$ is a separate training set, which we will not use to calibrate or test the trained CP algorithms.
Figure \ref{figure toy example 1} shows the label-space PIs obtained by setting $\theta = \theta_{flow}$ in $b_{flow}$ ($C_{flow}$, in red) and $\theta = \theta_{ER}$  in $b_{ER}$ ($C_{ER}$, in blue).

\section{Theory}
\label{section theory}
In this section, 
${\cal X}$ is an arbitrary attribute space and $\{ (X_{n}, Y_n) \in {\cal X} \times {\mathbb R}\}_{n=1}^{N+1}$ a collection of i.i.d. random variables from an unknown joint distribution, $P_{XY} = P_{Y|X} P_X$.
The regression model, $f(X_n) \approx {\rm E}(Y_n|X_n)$, $n=1, \dots, N + 1$, is assumed to be pre-trained on separate data.

\subsection{Quantiles}
\label{section quantiles}
Given a random variable, $Z \in {\cal Z}$ and its distribution, $P_{Z}$, let $F_{Z}(z) = P_{Z}(Z \leq z)$ be the Cumulative Distribution Function of $P_Z$. 
The $(1 - \alpha)$-th quantile of $Z \sim P_Z$ is 
\begin{align}
\label{theoretical quantile definition}
\bar Q_Z = \inf_q\{q \in {\cal Z}: F_Z(q) \geq (1 - \alpha) \}    
\end{align}
When $Z$ is continuous, $F_Z$ is strictly increasing and  $\bar Q_Z = F_Z^{-1}(1 - \alpha)$.
The $(1-\alpha)$-th sample quantile of a collection of i.i.d. random variables, $\{ Z_n \sim P_Z\}_{n=1}^N$, is the $(1-\alpha)$-th quantile of their empirical distribution $P_Z \approx N^{-1}\sum_{n=1}^N {\bf 1}(Z = Z_n)$, i.e. 
\begin{align}
\label{sample quantile definition}
    &Q_Z = \inf_q \{q \in {\cal Z}, \sum_{n=1}^N {\bf 1}(Z_n \leq q) \geq n_*\} \\
    &n_* = \lceil(N + 1)(1 - \alpha)\rceil\nonumber
\end{align}
where $\lceil s \rceil$ the smallest integer greater than or equal to $s \in {\mathbb R}$.
Assuming ties occur with probability 0, i.e. ${\rm Prob}(Z_n=Z_{n'})=0$ for any $n\neq n'$, $Q_Z$ is the  $n_*$-th smallest element of $\{ Z_n \sim P_Z\}_{n=1}^N$.
CP validity is a direct consequence of 
\begin{lemma}[Quantile Lemma \cite{tibshirani2019conformal}]
\label{lemma quantile}
Let $Z_1, \dots, Z_{N}, Z_{N+1} \in {\mathbb R}$ be a collection of i.i.d. random variables and $Q_Z$ be the $(1-\alpha)$-th sample quantile of $\{ Z_n \}_{n=1}^N$ defined in \eqref{sample quantile definition}. 
If ties occur with probability 0,  
\begin{align}
{\rm Prob} \left(Z_{N+1} \leq Q_Z \right) = \frac{\lceil (1 - \alpha) (N + 1)\rceil}{N +1}
\end{align}
\end{lemma}
The lemma first appeared in \cite{papadopoulos2002inductive}. 
Slightly different proofs can be found in  
\cite{lei2014distribution,tibshirani2019conformal, angelopoulos2021gentle}. 
The standard CP bounds, $1-\alpha \leq {\rm Prob}(Z_{N+1} \leq Q_Z) \leq 1-\alpha + \frac{1}{N+1}$, follows from $\lceil s \rceil - s\geq 0$ and $(1 - \alpha)(N+1) \leq \lceil (1-\alpha)(N+1)\rceil \leq (1 - \alpha) (N+1) + 1$. 
Asymptotically, $Q_Z$ is normally distributed around $\bar Q_Z$ with variance $\sigma^2 = \frac{(1-\alpha)\alpha}{N p_Z(\bar Q_Z)}$, where $p_Z(\bar Q_Z)$ is the density of $P_Z$ evaluated at $Z = \bar Q_Z$, with $\bar Q_Z$ defined in \eqref{theoretical quantile definition}. 

\subsection{Conformity scores}
A conformity score is a random variable, $A= a(f(X), Y)$, that describes the conformity between a prediction, $f(X)$, and the corresponding label, $Y$.
A standard choice is $a=|Y - f(X)|$. 
Let $P_{AX}$ be the distribution of the i.i.d. random variables $\{(A_{n}=|Y_n - f(X_n)|, X_{n})\}_{n=1}^{N+1}=$.
Lemma \ref{lemma quantile} guarantees the validity of the symmetric PI, 
\begin{align}
\label{marginal prediction intervals}
    C_A = [f(X_{N+1}) - Q_A,  f(X_{N+1}) + Q_A] 
\end{align}
when $Q_{A}$ is the $(1-\alpha)$-th sample quantile of $\{ A_n\}_{n=1}^N$.
We may also let the conformity scores be $B=b(A)$, where $b$ is a \emph{global monotonic function} of its argument, e.g.  $b(A)=-A^{-1}$ or $b(A) = \log A$.
In that case, we obtain the PIs by inverting $b$ and letting $C_{B} = [f(X_{N+1}) - b^{-1}(Q_{B}), f(X_{N+1}) + b^{-1}(Q_{B})]$, where $Q_{B}$ is the $(1-\alpha)$-th sample quantile of $\{ B_n=b(A_n)\}_{n=1}^N$ and $b^{-1}$ is defined by $b^{-1} \circ b(A) = A$.
For example, $b^{-1}(Q_{B}) = - \frac{1}{Q_{B}}$ if $B=-\frac{1}{A}$  and $b^{-1}(Q_{B})  = \exp(Q_{B})$ if $B = \log A$. 
Assuming ties occur with probability 0, $Q_{A}$ is the $n_*$-th smallest element of $\{A_n\}_{n=1}^N$, with $n_* = \lceil (1-\alpha) (N + 1)\rceil$.
Let $A_{*}$ be that element.
The $(1-\alpha)$-th sample quantile of the transformed scores, $Q_{B}$, is the $\lceil (1-\alpha) (N + 1)\rceil$-th smallest element of $\{b(A_n)\}_{n=1}^N$.
If $b$ is monotonic and applies globally to all samples, $b(A_n) < b(A_{n'})$ if and only if $A_n<A_{n'}$, for any $n\neq n'$.
Then $Q_{B} = b(A_{*})$ and $b^{-1}(Q_{B})= Q_{A}$, i.e. the size of the PIs does not depend on $b$.
If $b$ depends on the input, $b(A_n, X_n) < b(A_{n'}, X_{n'})$ does not imply $A_n<A_{n'}$, for any $n\neq n'$, i.e. the PIs depends on $b$.

\subsection{Normalizing Flows}
This work is about finding an input-dependent transformation $b =b(A, X)$ that changes the PIs to make them locally adaptive and more efficient automatically, i.e. without splitting the calibration data set and applying any existing CP algorithm.
In what follows, we assume $b$ always satisfies 
\begin{assumption}
    \label{assumption phi}
    For ${\cal A}, {\cal B} \subset {\mathbb R}$, $b: {\cal A} \times {\cal X} \to {\cal B}$
    \begin{enumerate}
    \item is strictly increasing on its first argument, i.e. $J_{b}(A, X) = \frac{\partial }{\partial A} b(A, X) > 0$ for all $(A, X)$ and
    \item its domain and co-domain are the same for all $X \in {\cal X}$.
    \end{enumerate}
\end{assumption}
Let $b^{-1}(B, X)$ be defined by $b^{-1}(b(A, X), X) = A$.
The assumption on the domain and co-domain of $b$ guarantees $b^{-1}(b(A, X'), X)$ is well defined for any $X \neq X'$.
We avoid over-fitting by letting $b$ be smooth in $X$ and $A$.
Since $b$ acts on random variables and obeys Assumption \ref{assumption phi}, we can interpret it as (part of) an NF. 
Let $P_Z$ and $U_Z$ be two distributions with the same support, ${\cal Z}$. 
An NF is an invertible coordinate transformation from ${\cal Z}$ to ${\cal Z}$ such that 
\begin{align}
    Z' = \phi_b(Z)  \sim U_{Z'},  \quad Z = \phi_b^{-1}(Z') \sim P_Z
\end{align}
In our case, $Z = (A, X)$, $Z'=(B, X)$, and $\phi_b(A, X) = (b(A, X), X)$. 
The Jacobian of $\phi_b$ is a $(|{\cal X}|+1)$-dimensional squared matrix, $J_{\phi_b}$, such that $J_{\phi_b ij} = 0$ for all $i, j > 1$ and $i\neq j$, $J_{\phi_b ii} = 1$ for all $i> 1$, $J_{\phi_b 1 i} = \frac{\partial}{\partial X_i} b(A, X)$ for all $i>1$, and $J_{\phi_b 11} = \frac{\partial}{\partial A} b(A, X)$.
We often use $J_b(A, X)$ instead of $J_{\phi_b 11}$.
Assumption \ref{assumption phi} implies $J_{\phi_b 11}> 0$ and guarantees the invertibility of $\phi_b$ because,
for any $(A, X)$, ${\rm det} (J_{\phi_b}(A, X)) = \prod_{i=1}^{|{\cal X}|+1} J_{\phi_b \ ii}(A,X) = J_{\phi_b \ 11}(A, X)$ is strictly positive.
When not explicitly required, we drop the trivial part of $\phi_{b}$ and use $b$ fo $\phi_b$ \emph{and} $\phi_{b1}$ depending on the context.
See \cite{papamakarios2021normalizing} for a review of using NFs in inference tasks. 

\subsection{Validity}
Given an NF, $b$, we let the associated marginal PI at $X_{N+1}$ be  
\begin{align}
\label{prediction interval phi}
    &C_{B} = [f(X_{N+1})- \delta, f(X_{N+1})+ \delta] \\ 
    &\delta = b^{-1}(Q_B, X_{N+1}) \nonumber 
\end{align}
where $Q_{B}$ is the $(1-\alpha)$-th sample quantile of $\{ B_n= b(A_n, X_n) \}_{n=1}^N$.
If ties occur with probability 0, the validity of $C_B$ defined in \eqref{prediction interval phi} is guaranteed by  
\begin{lemma}
\label{lemma validity cphi}
    Let $b$ satisfy Assumption \ref{assumption phi} and $C_B$ be the PI defined in \eqref{prediction interval phi}.
    Then 
    \begin{align}
        {\rm Prob}(Y_{N+1} \in C_B) =  \frac{\lceil (1-\alpha)(N+1)\rceil}{N+1}  
    \end{align}
\end{lemma}
The transformation is globally defined but \emph{acts differently} on the samples, e.g. we may have $b(A, X_n) \neq b(A, X_{n'})$ for some $A\in {\cal A}$ and $n\neq n'$.
The ranking of the \emph{original scores}, $\{A_n\}_{n=1}^N$, may differ from the ranking of the \emph{transformed scores}, $\{B_n\}_{n=1}^N$, i.e. $A_1<A_2< \dots< A_N$  may not imply $B_1<B_2< \dots< B_N$.
This happens if $A_n < A_{n'}$ and $b(A_n, X_n) > b(A_{n'}, X_{n'})$ for some $n\neq n'$. 
While validity is automatically guaranteed because calibration and test samples remain exchangeable, we may have $C_A \neq C_B$, e.g. when $b$ changes the ranking of the calibration samples.
Under further mild assumptions on $b$, Lemma  \ref{lemma size is different} shows that we can find a test object for which the PIs obtained with $b$ and $a$ have difference sizes, i.e. $|C_B| \neq |C_A|$. 
\begin{lemma}
\label{lemma size is different}
Let $\{ A_{n}\}_{n=1}^{N+1}$ be a collection of i.i.d. continuous random variables.
Assume $b$ satisfies Assumption \ref{assumption phi}.
Then, if $b(A_n, X_{N+1}) \neq b(A_n, X_n)$ for any $n=1, \dots, N$,
\begin{align}
    |C_{B}| \neq |C_A|
\end{align}
with $C_B$ and $C_A$ defined in \eqref{prediction interval phi} and \eqref{marginal prediction intervals}.
\end{lemma}

\subsection{Exact Normalizing Flows}
In some cases, marginally valid PIs are also conditionally valid for any $X_{N+1} \in {\cal X}$, i.e. $C_A$ defined in \eqref{marginal prediction intervals} obeys
\begin{align}
\label{special case}
    {\rm Prob}(Y_{N+1} \in C_{A} |X_{N+1}) \geq 1 - \alpha
\end{align}
This may occur when $P_{AX}$ has a specific form.
When the data are \emph{not} heteroscedastic, i.e. $P_{AX} =  P_{A|X} P_X = P_A P_X$, the equivalence of marginal and conditional PIs is guaranteed by 
\begin{theorem}
    \label{theorem marginal and conditional}
    Let $P_{AX} =  P_A P_X$ for any $X \in {\cal X}$.
    For any $X_{N+1} \in {\cal X}$, $C_A$ defined in \eqref{marginal prediction intervals}, obeys 
    \begin{align}
    \label{claim marginal and conditional}
        {\rm Prob}(Y_{N+1}\leq C_A|X_{N+1}) = \frac{\lceil (N+1) (1 - \alpha) \rceil}{N+1}
    \end{align}
   \end{theorem}
Theorem \ref{theorem marginal and conditional} is a straightforward consequence of the Bayesian theorem and Lemma \ref{lemma quantile}.
We include it here because it suggests we can find an NF that localizes the PIs.
The idea is to train $b$ to make $C_{B} = C_{b(A)}$ conditionally valid through Theorem \ref{theorem marginal and conditional}, i.e. to make $b$ such that $(b(A_n, X_n), X_n) = (B_n, X_n) \sim P_{BX} = P_B P_X$.
Interpreting $b$ as an NF, we can find a near-optimal $b$ through standard NF-training techniques, e.g. by \emph{maximizing the likelihood of the transformed scores under an arbitrary target distribution, $U_{B}$, that does not depend on the input}. 
Given samples from $A$, we need the composition between the target distribution and the score transformation, $b$.
In particular, $\int_{x}^{x'} dx p(f(x)) = \int_{f(x)}^{f(x')} \frac{dy}{f'(f^{-1}(y))} p(y)$ implies the density of the composition is $p(B, X) = u(b(A, X)) J_{b}(A, X)) p(X)$.
The objective function is  
\begin{align}
\label{likelihood}
\ell(b) &=  {\rm E} \left(\log u(B)p(X)  \right)\\
&= {\rm E}\left( \log\left(u(b(A, X)) \ |J_b(A, X)|\right) \right) + \ell_0 \nonumber 
\end{align}
where $u$ is the density of the (arbitrary) target distribution, $U_B$, and $\ell_0 = {\rm E} (\log p(X))$ does not depend on $b$.
Fix a given target distribution, $U_B$, e.g. let $U_B$ be the univariate Gauss distribution or $U_B\sim {\rm Uniform}([0, 1])$.
Assume there exists an NF, $b$, that satisfies Assumption \ref{assumption phi} and is such that $P_{BX} = U_{B} P_X$ for any $(A, X)$ when $B=b(A, X)$.
Then, $C_B$ defined in \eqref{prediction interval phi} is conditionally valid at $X_{N+1}$. The claim is supported by 
\begin{corollary}
    \label{corollary exact flow intervals}
    Let $U_B$ be an arbitrary univariate distribution and $b$ an NF satisfying Assumption \ref{assumption phi}.
    If $(B, X) = (b(A, X), X) \sim P_{BX} = U_B P_X$ for any $(A, X)$, $C_B$ defined in \eqref{prediction interval phi} obeys
    \begin{align}
        {\rm Prob}(Y_{N+1} \in C_{B}|X_{N+1}) = \frac{\lceil(1 - \alpha) (N + 1) \rceil}{N+1}
    \end{align}
\end{corollary}
Corollary \ref{corollary exact flow intervals} follows from Lemma \ref{lemma validity cphi} and the monotonicity of $b$.
There is no contradiction with the negative results of \cite{lei2012distribution, vovk2012conditional} because exact factorization can not be achieved with finite data. 

\subsection{Non-exact Normalizing Flows}
Let $\hat b$ be an NF trained by maximizing a finite-sample empirical estimation of the likelihood defined in \eqref{likelihood}.
We do not expect $\hat b$ to factorize $P_{BX}$ exactly but assume it approximates the ideal optimal transformation, $b$, defined in Corollary \ref{corollary exact flow intervals} in the Huber sense.
More precisely, we let $\epsilon >0$ quantify the discrepancy between the two transformations and
\begin{align}
\label{huber expansion}
    &\hat b = (1 - \epsilon) b + \epsilon \delta, 
    \end{align}
where $\delta = \delta(A, X)$ is an unknown error term that depends on $(A, X)$.
The assumption is technical and used to prove the error bounds below. 
The density of the perturbed distribution is $p(\hat B, X) = |J_{\hat b}(A, X)| u(\hat b(A, X)))p(X)$, 
which may be expanded in $\epsilon$ under the assumption $\epsilon <<1$. 
Theorem \ref{theorem approximate validity} characterizes the validity of $C_{\hat B}$, i.e. the PIs defined in \eqref{prediction interval phi} with $b$ replaced by $\hat b$, up to $o(\epsilon^2)$ errors.
We assume $b$ and $\hat b$ fulfill the requirements of Assumption \ref{assumption phi}, $b$ satisfies the assumption of Corollary \ref{corollary exact flow intervals}, and $\hat b$ is the minimizer of \eqref{likelihood} for a given target distribution $U_B$.
To simplify the notation, we let $B=b_X(A)$ where $b_X = b(A, X)$ (idem $b_{X}^{-1}$,  $\hat b_X$, and $\hat b^{-1}_X$) and define $\tilde B = \psi_X(A)$, where $\psi_{X} = b_{X_{N+1}}^{-1} \circ \hat b_{X_{N+1}}\circ \hat b_{X}^{-1} \circ b_{X}$.  
We bound the validity gap of $C_{\hat B}$ in terms of the variation distance between the distributions of $B$ and $\tilde B$, i.e. 
\begin{align}
d_{\rm TV}(P_{BX}, P_{\tilde  BX})  &=  \sup_{(A, X)} \| p(B, X) - p(\tilde B, X) \| 
\end{align}
where $p(B, X) $ and $p(\tilde B, X) $ are the densities of $P_{BX} = U_B P_X$ and $P_{\tilde BX}$ and $B$ and $\tilde B$ depend on $(A, X)$ through $b$ and $\psi$.
We use the Maximal Coupling Theorem to link the CP validity bound in \eqref{lemma validity cphi} and the total variation distance above. 
See \cite{lindvall2002lectures} or \cite{ross2023second} for an overview of coupling methods.
Up to  $o(\epsilon^2)$ corrections, an explicit lower bound of the gap is given in 
\begin{theorem}
\label{theorem approximate validity}
Let $b(A, X)$ and $\hat b(A, X)$ obey Assumption \ref{assumption phi} and $U_B = {\rm Uniform}([0, 1])$.
Assume $\hat b$ obeys \eqref{huber expansion} for all $(A,X)$. 
Then,   
\begin{align}
    &{\rm Prob}(B_{N+1} \leq Q_{\hat B}) \\
    &\geq \frac{\lceil (N+1)(1 - \alpha)\rceil }{N+1} - \frac{1}{2} d_{\rm TV}(P_B, P_{\tilde B}) \\
    &\geq \frac{\lceil (N+1)(1 - \alpha)\rceil }{N+1} - \epsilon \sup_{x} \|p_X(x) \| L_\delta L_{b^{-1}} + o(\epsilon^2) \nonumber 
\end{align}
where $Q_{\hat B}$ is the sample quantile of $ \{\hat b(A_n, X_n)\}_{n=1}^N$ defined in \eqref{sample quantile definition} with $b$ replaced by $\hat b$, $\tilde B = \psi_X(B)$, $\psi_{X} = b_{X_{N+1}}^{-1} \circ \hat b_{X_{N+1}}\circ \hat b_{X}^{-1} \circ b_{X}$, $b_{X}(A) = b(A, X)$ (idem $b_{X}^{-1}$,  $\hat b_X$, and $\hat b^{-1}_X$), $L_\delta$ and $L_{b^{-1}}$ are the Lipschitz constants of $\delta(B, X)$ and  $b^{-1}$, and $p(X)$ is the marginal density of the covariates.
\end{theorem}
Theorem \ref{theorem approximate validity} connects our work with the non-exchangeability gaps obtained in  \cite{barber2022conformal} in a different framework.

\section{Implementation}
\label{section implementation}
We compare two models trained with the proposed scheme, a standard CP algorithm, and the ER model of \cite{papadopoulos2008normalized}. 
For simplicity, we focus on Split CP, where the regressor, $f$, is pre-trained on separate data and kept fixed.

\subsection{Data}
We generate 4 synthetic data sets by perturbing the output of a polynomial regression model of order 2 with four types of heteroscedastic noise.
Each data sets consist of 1000 samples of a pair of random variables, $(X, Y)$, obeying
\begin{align}
&Y = X^T w + \epsilon_i,\\
    &X = [1, X_1, X_1^2], \quad X_1 \sim {\rm Uniform}([-1, 1]), \nonumber \\
    &\epsilon_i = 0.1 + \sigma_{\tt synth-i}(X) E, \quad E \sim {\cal N}(0, 1) \nonumber
\end{align}
where $w \in {\mathbb R}^3$ is a randomly generated fixed parameter,  ${\tt i} \in \{ {\tt cos}, {\tt squared}, {\tt inverse}, {\tt linear} \}$, and 
\begin{align}
    &\sigma_{\tt synth-cos}(X) = 2 \cos\left(\frac{\pi}{2} X_1 \right) {\bf 1}(X_1 < 0.5) \\
    &\sigma_{\tt synth-squared}(X) = 2 X^2_1 {\bf 1}(X_1 > 0.5)\\
    &\sigma_{\tt synth-inverse}(X) = 2 \frac{1}{0.1 + |X_1|}{\bf 1}(X_1 < 0.5)\\
    &\sigma_{\tt synth-linear}(X) = 2 |X_1|{\bf 1}(X_1 > 0.5)
\end{align}

For the real-data experiments, we use the following 6 public benchmark data sets from the UCI database:
{\tt bike}, the Bike Sharing Data Set \citep{bikeData}, {\tt CASP}, the Physicochemical Properties of Protein Tertiary Structure Data Set \citep{caspData}, {\tt community}, Community and Crime Data Set\citep{communityData}, {\tt concrete}, the Concrete Compressive Strength Data Set \citep{concreteData}, {\tt energy}, the Energy Efficiency Data Set \citep{energyData}, and {\tt facebook\_1}, the Facebook Comment Volume Data Set \cite{facebookData}. 

All data sets are split into two subsets.
We use the first subset to train a Random Forest (RF) regressor and the second subset to train and test the conformity functions.
For stability, we limit the attribute dimensions to 10 (with PCA) and normalize the label before training the RF models.
The Mean Absolute Error of the RF regressor is reported in Table \ref{table mae}.
%%%
\begin{table}
    \centering
    \begin{small}
    \begin{tabular}{*{2}{l}} 
    \toprule
    data set & MAE \\ 
    \midrule
{\tt synth-cos}& 0.051(0.033)\\
{\tt synth-inverse}& 0.056(0.007)\\ 
{\tt synth-linear}& 0.157(0.054)\\
{\tt synth-squared} &0.125(0.038)\\
{\tt bike}& 0.028(0.001)\\ 
{\tt CASP}& 0.14(0.003)\\ 
{\tt community}& 0.007(0.001)\\ 
{\tt concrete}& 0.051(0.002)\\ 
{\tt energy}&0.019(0.001)\\ 
{\tt  facebook\_1}&0.003(0.0)\\
\bottomrule
    \end{tabular}
    \caption{Averages and standard deviation over 5 runs of the MAE of the RF regression model on the synthetic and real data sets.}
    \label{table mae}
    \end{small}
\end{table}
To make the performance comparable across different data sets, we reduce the size of the second subset to 1000 (except for {\tt community}, {\tt concrete}, and {\tt energy} that have size 997, 515, and 384), split it into two equal parts, and use the first to train the conformity measures and the second for calibrating and testing the optimized models.

\subsection{Models}
We let $A = |Y - f(X)|$ and consider four model classes,
\begin{align}
    &b_{\tt baseline} = A \\
    &b_{\tt ER} = \frac{A}{\gamma + |g(X)|} \\
    &b_{\tt Gauss} = \log\left(\frac{A}{\gamma + |g(X)|}\right) \\
    &b_{\tt Uniform} = \sigma\left(\frac{A}{\gamma + |g(X)|}\right) 
\end{align}
where $\gamma = 0.001$ and $g$ is a fully connected ReLU neural network with 5 layers of 100 hidden units per layer. 
The network parameters of {\tt ER} are trained as in \cite{papadopoulos2008normalized} by minimizing $\ell_{ER} = {\rm E}(|g(X)|-|f(X)-Y|)^2$. 
{\tt Gauss} and {\tt Uniform} are trained with the proposed approach by maximizing the log-likelihood in \eqref{likelihood} where $U_B = {\cal N}(0, 1)$ for {\tt Gauss} and $U_B = {\rm Uniform}([0, 1])$ for {\tt Uniform}.
The model functional form guarantees $b$ belongs to the distribution support for any $(A, X)$.
We use the ADAM gradient descent algorithm of \cite{adam1904understanding} to solve all optimization problems with standard parameters.
The learning rate is 0.01 for {\tt ER}, $10^{-4}$ for {\tt Gauss}, and $10^{-5}$ for {\tt Uniform}.

\subsection{Results}
To evaluate the PIs, we consider their empirical validity, ${\rm E}({\bf 1}(Y_{N+1} \in C_B))$, average size, ${\rm E}(|C_B|)$, and empirical input-conditional coverage, which we approximate with the Worse-Slab Coverage (WSC) algorithm of \cite{cauchois2020knowing}. % 
Table \ref{table all exps} summarizes our numerical results across the 4 synthetic and 6 real data sets for three values of the confidence level, $1-\alpha \in \{ 0.95, 0.90, 0.65 \}$. 
Tables \ref{table data sets synth} and \ref{table data sets real} show the model performances at $\alpha = 0.05$ on each data set. 
The figures are the averages and standard deviations over 5 random train-test splits.

{\tt baseline} is the best method for $\alpha = 0.35$, on synthetic and real data, but is generally outperformed by the trained models at higher confidence levels.
{\tt Gauss} seems to outperform all other models on synthetic data. 
This may be due to the Gaussianity of the noise in the generation of the synthetic samples. 
On synthetic data, {\tt ER} is the second best model, probably because we generate the data using $Y \sim f(X) + g(X) \epsilon$, $\epsilon \sim {\cal N}(0, 1)$, which implies the ER assumptions are \emph{exact}.
{\tt Uniform} is the best model on real data. 
Interestingly, the model with the conditional coverage closest to the nominal is not the same at all confidence levels.
Table \ref{table data sets real} suggests that the optimized models are outperformed by {\tt baseline} when data are not heteroscedastic, i.e. when {\tt baseline} has good conditional coverage.
This seems to be a shared problem of reweighting methods, as already observed in \cite{romano2019conformalized}.
The code for reproducing all numerical simulations is available in this \href{https://github.com/nicoloRHUL/NormalizingFlowsForConformalRegression}{gitHub repository}. 
\begin{table}
    \centering
    \begin{small}
    \begin{tabular}{*{4}{l}}
\toprule
\multicolumn{4}{c}{{\bf synthetic data} (all data sets)}\\ 
&coverage&size&WSC\\ \midrule
\multicolumn{4}{c}{$\alpha=0.05$} \\ 
baseline                &0.954(0.021)   &0.783(0.086)   &0.798(0.198)\\
ER              &0.954(0.019)   &0.57(0.122)    &0.915(0.094)\\
Gauss           &0.953(0.02)    &0.506(0.048)   &0.883(0.118)\\
Uniform         &0.955(0.025)   &0.624(0.077)   &0.881(0.131)\\
\midrule
\multicolumn{4}{c}{$\alpha=0.1$} \\ 
baseline                &0.904(0.035)   &0.546(0.063)   &0.637(0.239)\\
ER              &0.902(0.027)   &0.429(0.067)   &0.833(0.147)\\
Gauss           &0.904(0.024)   &0.382(0.035)   &0.8(0.119)\\
Uniform         &0.906(0.036)   &0.451(0.041)   &0.737(0.153)\\
\midrule
\multicolumn{4}{c}{$\alpha=0.35$} \\ 
baseline                &0.661(0.042)   &0.183(0.018)   &0.244(0.146)\\
ER              &0.677(0.052)   &0.209(0.029)   &0.443(0.134)\\
Gauss           &0.673(0.05)    &0.2(0.022)     &0.461(0.139)\\
Uniform         &0.668(0.05)    &0.22(0.029)    &0.449(0.13)\\
\bottomrule
\vspace{0.5cm}
\\\toprule
\multicolumn{4}{c}{{\bf real data} (all data sets)}\\ 
&coverage&size&WSC\\ \midrule
\multicolumn{4}{c}{$\alpha=0.05$} \\ 
baseline                &0.955(0.016)   &0.141(0.013)   &0.929(0.066)\\
ER              &0.955(0.02)    &0.175(0.056)   &0.94(0.079)\\
Gauss           &0.957(0.013)   &0.17(0.03)     &0.913(0.078)\\
Uniform         &0.952(0.017)   &0.137(0.015)   &0.944(0.076)\\
\midrule
\multicolumn{4}{c}{$\alpha=0.1$} \\ 
baseline                &0.901(0.029)   &0.103(0.009)   &0.853(0.134)\\
ER              &0.9(0.032)     &0.117(0.024)   &0.832(0.114)\\
Gauss           &0.911(0.025)   &0.113(0.011)   &0.889(0.093)\\
Uniform         &0.901(0.028)   &0.102(0.009)   &0.869(0.11)\\
\midrule
\multicolumn{4}{c}{$\alpha=0.35$} \\ 
baseline                &0.681(0.044)   &0.045(0.005)   &0.512(0.157)\\
ER              &0.659(0.039)   &0.049(0.006)   &0.645(0.143)\\
Gauss           &0.658(0.041)   &0.048(0.003)   &0.574(0.144)\\
Uniform         &0.672(0.054)   &0.046(0.005)   &0.643(0.122)\\
\bottomrule
    \end{tabular}
    \caption{Average efficiency of the model PIs (coverage, size, and Worst Slab Coverage (WSC) estimate of the conditional coverage \citep{cauchois2020knowing}) over the data sets listed in Table \ref{table mae}.
    The reported averages and standard deviation are computed over 5 random training-test splits.}
    \label{table all exps}
    \end{small}
\end{table}

\begin{table}
    \centering
    \begin{small}
    \begin{tabular}{*{4}{l}}
\toprule
\multicolumn{4}{c}{\bf synthetic data  ($\alpha = 0.05$)}\\
&coverage&size&WSC\\ \midrule
\multicolumn{4}{c}{\tt synth-cos} \\ 
%--------------------------------------
baseline                &0.944(0.029)   &0.274(0.039)   &0.749(0.293)\\
ER              &0.945(0.022)   &0.208(0.123)   &0.926(0.062)\\
Gauss           &0.945(0.024)   &0.128(0.015)   &0.833(0.136)\\
Uniform         &0.945(0.035)   &0.148(0.03)    &0.828(0.188)\\
\midrule
\multicolumn{4}{c}{\tt synth-inverse} \\ 
%+++++++++++++++++++++++synth-inverse
baseline                &0.953(0.02)    &0.321(0.048)   &0.758(0.231)\\
ER              &0.96(0.017)    &0.165(0.019)   &0.942(0.064)\\
Gauss           &0.944(0.024)   &0.124(0.021)   &0.787(0.225)\\
Uniform         &0.951(0.021)   &0.195(0.024)   &0.814(0.233)\\
\midrule
\multicolumn{4}{c}{\tt synth-linear} \\
%+++++++++++++++++++++++synth-linear
baseline                &0.967(0.018)   &1.923(0.157)   &0.826(0.131)\\
ER              &0.96(0.011)    &1.54(0.309)    &0.941(0.063)\\
Gauss           &0.967(0.008)   &1.402(0.109)   &0.927(0.09)\\
Uniform         &0.968(0.014)   &1.795(0.201)   &0.898(0.071)\\
\midrule
%+++++++++++++++++++++++synth-squared
\multicolumn{4}{c}{\tt synth-squared} \\ 
baseline                &0.953(0.016)   &0.614(0.099)   &0.858(0.137)\\
ER              &0.949(0.026)   &0.366(0.037)   &0.849(0.185)\\
Gauss           &0.956(0.024)   &0.37(0.048)    &0.987(0.018)\\
Uniform         &0.956(0.029)   &0.359(0.054)   &0.983(0.03)\\
\bottomrule
\end{tabular}
    \caption{Efficiency of the model PIs at $\alpha=0.05$ on the synthetic data sets listed in Table \ref{table mae}.
    The reported averages and standard deviation are computed over 5 random training-test splits.}
    \label{table data sets synth}
\end{small}
\end{table}
\begin{table}
    \centering
    \begin{small}
    \begin{tabular}{*{4}{l}}
\\\toprule
\multicolumn{4}{c}{\bf real data ($\alpha = 0.05$)}\\
&coverage&size&WSC\\ \midrule
\multicolumn{4}{c}{\tt bike} \\ 
%+++++++++++++++++++++++bike
baseline                &0.98(0.011)    &0.112(0.011)   &0.995(0.011)\\
ER              &0.969(0.019)   &0.181(0.095)   &0.995(0.011)\\
Gauss           &0.972(0.007)   &0.127(0.026)   &0.98(0.017)\\
Uniform         &0.976(0.012)   &0.119(0.021)   &0.982(0.02)\\
\midrule
\multicolumn{4}{c}{\tt CASP} \\ 
%+++++++++++++++++++++++CASP
baseline                &0.944(0.021)   &0.413(0.026)   &0.969(0.041)\\
ER              &0.959(0.011)   &0.435(0.036)   &0.883(0.192)\\
Gauss           &0.973(0.009)   &0.525(0.063)   &0.959(0.045)\\
Uniform         &0.955(0.011)   &0.433(0.03)    &0.89(0.196)\\
\midrule
\multicolumn{4}{c}{\tt community} \\ 
%+++++++++++++++++++++++ community
baseline                &0.965(0.011)   &0.059(0.018)   &0.888(0.097)\\
ER              &0.967(0.009)   &0.035(0.016)   &0.967(0.02)\\
Gauss           &0.968(0.011)   &0.071(0.028)   &0.924(0.06)\\
Uniform         &0.963(0.013)   &0.03(0.007)    &0.958(0.052)\\
\midrule
%
%+++++++++++++++++++++++concrete
\multicolumn{4}{c}{\tt concrete} \\ 
baseline                &0.96(0.023)    &0.154(0.016)   &0.938(0.064)\\
ER              &0.947(0.031)   &0.288(0.134)   &0.948(0.078)\\
Gauss           &0.95(0.016)    &0.201(0.045)   &0.954(0.043)\\
Uniform         &0.958(0.021)   &0.159(0.014)   &0.972(0.039)\\
\midrule
%+++++++++++++++++++++++energy
\multicolumn{4}{c}{\tt energy} \\ 
baseline                &0.942(0.016)   &0.092(0.005)   &0.965(0.046)\\
ER              &0.944(0.041)   &0.098(0.05)    &0.897(0.131)\\
Gauss           &0.944(0.017)   &0.079(0.015)   &0.793(0.18)\\
Uniform         &0.933(0.028)   &0.071(0.019)   &0.975(0.033)\\
\midrule
%
%+++++++++++++++++++++++facebook_1
\multicolumn{4}{c}{\tt facebook\_1} \\ 
baseline                &0.94(0.014)    &0.014(0.003)   &0.82(0.139)\\
ER              &0.947(0.011)   &0.016(0.005)   &0.953(0.045)\\
Gauss           &0.935(0.02)    &0.015(0.004)   &0.87(0.126)\\
Uniform         &0.929(0.015)   &0.01(0.002)    &0.885(0.114)\\
\bottomrule
\end{tabular}
    \caption{Efficiency of the model PIs at $\alpha=0.05$ on the real data sets listed in Table \ref{table mae}.
    The reported averages and standard deviation are computed over 5 random training-test splits.}
    \label{table data sets real}
    \end{small}
\end{table}

\section{Related work}
\label{section related work}
%%%%%%%%%%%%%%%%
{\bf Calibration training}
In CP, learning a conformity function from data is fairly new. 
To the best of our knowledge, the only example of a trained conformity function is the ER algorithm of \cite{papadopoulos2008normalized, papadopoulos2011regression, lei2012distribution}, where localization is achieved by reweighting $|Y - f(X)|$ with a pre-trained model of the conditional residual, $|g(X)| \approx {\rm E}(|Y-f(X)|\ |X)$. 
Outside CP, there are several examples of calibration optimization for data science applications \citep{platt1999probabilistic, zadrozny2002transforming,naeini2015obtaining}.
See \cite{guo2017calibration} for an introduction and empirical comparison of different calibration methods for neural networks. 
{\bf Object-dependent conformity measures.}  \cite{papadopoulos2008normalized, papadopoulos2011regression, lei2012distribution} use different versions of reweighted conformity measures.
The localization function is either fixed, e.g. a KNN-based variance estimator, or pre-trained using \emph{ad-hoc} strategies. 
Section 5 of \cite{romano2019conformalized} contains a detailed discussion on the limitations of ER.
Despite its intuitive and empirical efficiency, ER has been poorly investigated or justified from a theoretical perspective.
Our work provides a conceptual framework to explain why it works well for approximating conditional validity  \citep{lei2012distribution, foygel2021limits}.
Recent work about ER includes \cite{vovk2020conformal}, which is a theoretical study on the validity of oracle conformity measures, and \cite{bellotti2021approximation}, where the conformity score is iteratively updated to make the PI conditionally valid. 
Similar to \cite{gibbs2021adaptive}, coverage is corrected by minimizing an empirical estimation of the validity gap.
Besides \cite{papadopoulos2008normalized, bellotti2020constructing}, conformity scores other than $A = |f(X)-Y|$ have been rarely used.
In \cite{romano2019conformalized}, the conformity function is redesigned to mimic the pinball loss of quantile regression problems.
In \cite{colombo2023training}, a series of trained conformity functions are tested empirically.
Compared to this work, the learning scheme is not analyzed theoretically and uses a different learning loss.
We are unaware of other works where the conformity measure is explicitly optimized.
{\bf CP localization and conditional validity.}
Except for \cite{papadopoulos2008normalized}, the scheme can be combined with other localization methods because it applies to any base conformity score.
\cite{papadopoulos2008normalized} is an exception because the conformity function is trained by minimizing ${\rm E}_{XY}|A^2 - g^2(X)|^2$. 
In \cite{lei2014distribution, vovk2012conditional, lin2021locally, guan2023localized, deutschmann2023adaptive}, locally adaptive PI are constructed by reweighting the calibration samples and temporarily breaking data exchangeability.
The weights transform the marginal distribution into an estimate of the object-conditional distribution.
Often, computing the localizing weights requires a density estimation step based on one or more hyper-parameters \citep{lei2014distribution, vovk2012conditional, guan2023localized, deutschmann2023adaptive}.  
This may cause technical issues and can be unreliable if data is scarce.
Our approach avoids an explicit estimation, because $b$ is a globally defined functional, and does not require splitting the calibration set.
Conditional validity gaps can be viewed as a non-exchangeability problem.
\cite{barber2022conformal} is a study of CP under general non-exchangeability but does not make an explicit connection to local adaptivity.
\cite{xu2023sequential} exploits the bounds of \cite{barber2022conformal} for proving the asymptotic convergence of the estimated PIs to the exact conditional PIs.  
Theorem 4 in \cite{guan2023localized} guarantees exact conditional coverage for a sample reweighting method, up to corrections on the estimated PI.
The NF setup allows more explicit bounds on the validity of the algorithm outputs (Theorem \ref{theorem approximate validity} in Section \ref{section theory}).
In \cite{einbinder2022training}, a point-prediction model is trained to guarantee $P_{AX}=U_A P_X$, where $U_A={\rm Uniform}([0, 1])$.
It is unclear whether tuning the point-prediction model or the conformity function produces equivalent PIs.
This work is intuitively close to conformity-aware training, which aims to optimize the output of a standard CP algorithm by tuning the underlying model \citep{colombo2020training, bellotti2020constructing, stutz2021learning, einbinder2022training}. 
The two ideas are compatible and could be implemented simultaneously. 
We leave this for future work.

\section{Discussion and limitations}
\label{section discussion and limitations}
This is mainly a theoretical and methodological work.
We recognize our numerical simulations are limited, especially regarding the model complexity. 
We also miss a full comparison with existing localization approaches. 
We focus on conformity functions similar to ER to underline the efficiency of the learning strategy, without bias coming from the definition of more or less suitable model classes.
Generalizing the approach to more complex NF is possible, provided $b(A, X)$ remains invertible, i.e. monotonic in $A$.
A comparison with other localization methods goes beyond our scope because calibration training is orthogonal to many existing strategies, e.g. algorithms based on reweighting the calibration samples.
The proposed scheme could be used on top of them to provide theoretical guarantees. 
As mentioned in Section \ref{section related work}, CP-aware retraining of the prediction model could also be combined with calibration training.

%%%%%%%%%%%%%%%%%%%%%%%%%%%%%%%%%%%%%%%%%%%%%%%%%%%%%%%%%%%%%%%%%%%%%%
\newpage
\bibliography{refs.bib}

%\newpage
%\onecolumn
%%\begin{center}
%\large{{\bf Normalizing Flows for Conformal Regression} \\
%Supplementary Material }    
%\end{center}
%\appendix
\section{Proofs}

%%%%%%%%%%%%%%%%%%%%%%%%%%%%%%%%%%%%%%%%%%%%%%%%
\paragraph{Proof of Lemma \ref{lemma quantile}}.
Assume ties occur with probability 0.
According to \eqref{sample quantile definition}, $Q_Z$ is the $n* = \lceil (1-\alpha)(N+1)\rceil$-th smallest element of $\{ Z_n \}_{n=1}^N$.
Assume the calibration samples have been labeled so that $Z_1 < Z_2 \dots < Z_{N-1} < Z_N$.
By assumption, $Z_1, \dots, Z_n$, and $Z_{N+1}$ are exchangeable.
This implies $Z_{N+1}$ falls with equal probability in any of the $N+1$ intervals 
\begin{align}
    (-\infty, Z_1), &[Z_1,Z_2) \dots, [Z_{n*-1}, Q_Z), \nonumber\\&(Q_Z, Z_{n*+1})\dots %\\ \dots  
    (Z_{N-1}, Z_{N}), [Z_{N}, \infty)
\end{align}
i.e.
%\begin{align}
${\rm Prob}(Z_{N+1} \leq Q_Z) = \frac{n*}{N+1} = \frac{\lceil (1-\alpha)(N+1)\rceil}{N+1}.$
%\end{align}
$\square$

\paragraph{Proof of Lemma \ref{lemma validity cphi}}
$\{ B_n\}_{n=1}^N$ are i.i.d. random variable because $b$ is deterministic and  $\{ A_n\}_{n=1}^N$ are i.i.d.
When $b$ satisfies Assumption \ref{assumption phi}, ${\rm Prob}(A_n=A_{n'}) = 0$ for any $n\neq n'$ implies 
${\rm Prob}(B_n=B_{n'})= {\rm Prob}(A_n = b^{-1}(b(A_{n'},X_{n'}), X_n) = 0$ for any $n\neq n'$, i.e. there are no ties in $\{ B_n\}_{n=1}^N$.
Let $Q_{B}$ be the $(1-\alpha)$-th sample quantile $\{ B_n\}_{n=1}^N$ defined in \eqref{sample quantile definition}.
From Lemma \ref{lemma quantile}, ${\rm Prob}(B_{N+1} \leq Q_{B}) = \frac{n_*}{N+1}$, with $n_* = \lceil (1-\alpha)(N+1)\rceil$.
Let $b_X(A) = b(A, X)$ and $b^{-1}_X(B) = b^{-1}(B, X)$, with $b^{-1}$ defined by $b(b^{-1}(B, X), X) = b_X \circ b_X^{-1}(B)$, and
$\partial_A b_X(A) = J_{b\ 1 1}(A, X)= \frac{\partial}{\partial A}b(A, X) = \frac{\partial}{\partial A}b_X(A)$.
By Assumption \ref{assumption phi},  $\partial_A b_X>0$ for all $X$.
Let $\frac{d}{ds} h(s, g(s)) = \partial_s h + \partial_g h \partial_s g$ be the total derivative of $h$.
From $1 = \frac{d}{dB} b_X\circ b_X^{-1}(B) = \partial_A b_X(b_X^{-1}(B)) \frac{d}{dB} b_X^{-1}(B)$, we obtain $\frac{d}{dB}b^{-1}_X(B) = \left( \partial_A b_X(b^{-1}_X(B))\right)^{-1} > 0$, i.e. $b^{-1}_{X_{N+1}}(B)$ is a monotonic function of $B$.
Therefore, 
\begin{align}
    &{\rm Prob}\left( B_{N+1} \leq Q_{B}\right)\\ 
    &={\rm Prob}\left(b^{-1}_{X_{N+1}}(B_{N+1}) \leq b_{X_{N+1}}^{-1}(Q_{B})\right) \\
    & ={\rm Prob}\left(b^{-1}_{X_{N+1}}\circ b_{X_{N+1}}(A_{N+1}) \leq b_{X_{N+1}}^{-1}(Q_{B})\right) \\
    & ={\rm Prob}\left(A_{N+1} \leq b_{X_{N+1}}^{-1}(Q_{B})\right) \\
    & ={\rm Prob}\left( |f(X_{N+1})-Y_{N+1}| \leq b_{X_{N+1}}^{-1}(Q_{B})\right) \\
    & ={\rm Prob}\left( Y_{N+1} \in C_B\right) 
\end{align}
where $C_B$ is defined in \eqref{marginal prediction intervals}.
$\square$
\paragraph{Proof of Lemma \ref{lemma size is different}}
Let $\{ B_n=b_{X_n}(A_n)\}_{n=1}^{N+1}$, where $b_X(A) = b(A, X)$, and $C_A$ and $C_B$ be the PIs in \eqref{marginal prediction intervals} and \eqref{prediction interval phi}.
From \eqref{sample quantile definition}, there are $m_*$ and $n_*$ such that $Q_A = A_{m*}$ and  $Q_{\hat B} = b_{X_{n_*}}(A_{n*})$.
Then, when $b_{X_{N+1}}(A_{n}) \neq  b_{X_{n}}(A_{n})$ for any $n$,  we have  $b_{X_{N+1}}(A_{n_*}) \neq  b_{X_{n_*}}(A_{n_*})$ and 
\begin{align}
        &|C_{B}| = b_{X_{N+1}}^{-1}\circ b_{X_{m_*}}(A_{m_*}) \neq  A_{n_*} =|C_A| 
\end{align}
The claim holds because $A_{n*}= b_{X_{N+1}}^{-1}\circ b_{X_{m_*}}(A_{m_*})$ occurs with probability 0 if $A_n$ are continuous.
$\square$
%%%
\paragraph{Proof of Theorem\ref{theorem marginal and conditional}}
    Let $\{A_n \sim P_{A} \}_{n=1}^{N}$ $\{ \tilde A_n \sim P_{A|X_{N+1}} \}_{n=1}^{N}$ be two collections of i.i.d random variables distributed according to the marginal and $X_{N+1}$-conditional distributions.
    Let $Q_{A}$ and $Q_{\tilde A}$ be the sample quantiles of the two collections defined in \eqref{sample quantile definition}.
    Let $C_{A}$ be the PI defined in \eqref{marginal prediction intervals} and $C_{\tilde A}$ be obtained analogously with $Q_A$ replaced by $Q_{\tilde A}$. 
    Assume ties occur with probability 0. 
    By the Bayesian theorem, $P_{AX} = P_A P_X$ implies $P_{A|X} = P_A = \sum_{X} P_{AX}$.
    Then, for any $X_{N+1}$, $\tilde A_n \sim P_A \sim A_n$ and, from Lemma \ref{lemma quantile},  ${\rm Prob}(\tilde A_{N+1} \leq Q_{\tilde A}) = {\rm Prob}(\tilde A_{N+1} \leq Q_{A}) = {\rm Prob}(Y_{N+1} \in C_A)$.
    $\square$
    %%%
\paragraph{Proof of Corollary \ref{corollary exact flow intervals}}
Let $Q_{A|X_{N+1}}$ and $C_{A|X_{N+1}}$ be the conditional sample quantile of $\{ \tilde A_n \sim P_{A|X_{N+1}} \}$ and the corresponding PI defined as in \eqref{marginal prediction intervals} with $Q_A$ replaced by $Q_{A|X_{N+1}}$. 
By construction, $C_{A|X_{N+1}}$ is conditionally valid at $X_{N+1}$, i.e. it obeys ${\rm Prob}(Y_{N+1} \in C_{A|X_{N+1}}|X_{N+1}) = \frac{m_*}{N+1}$, $m_*=\lceil(1 - \alpha) (N + 1) \rceil$. 
Let $(B, X) = (b(A, X), X) = (b_X(A), X)$.
Then, if $b$ obeys Assumption \ref{assumption phi} and $P_{BX} = P_{B|X} P_X = U_B P_X$, 
\begin{align}
    Q_{A|X_{N+1}} 
    &= Q_{b^{-1}_{X_{N+1}}(B)|X_{N+1}} \\
    &= b_{X_{N+1}}^{-1}(Q_{B|X_{N+1}}) = b_{X_{N+1}}^{-1}(Q_{B})\label{line3}
\end{align}
because $b^{-1}_{X_{N+1}}$ is monotonic and we apply it globally to all samples (second equality) and $P_{BX} =  U_B P_X$ implies $Q_{B|X_{N+1}} = Q_{B}$ (last equality).
The claim follows from Lemma \ref{lemma quantile}, $b_{X_{N+1}}(A)=b(A, X_{N+1})$, and the PI definition in \eqref{prediction interval phi}.
$\square$    
\paragraph{Proof of Theorem \ref{theorem approximate validity}.}
Let $\{ A_n\}_{n=1}^{N+1}$ be a collection of i.i.d. conformity scores and  $\{ B_n = b(A_n, X_n)\}_{n=1}^{N+1}$ and $\{ \hat B_n = \hat b(A_n, X_n)\}_{n=1}^{N+1}$ the conformity scores transformed by $b$ and $\hat b = (1 - \epsilon) b + \epsilon \delta$.
Let $C_B$ be the PI defined in \eqref{prediction interval phi} and $C_{\hat B}$  defined analogously by replacing $b$ with $\hat b$.
Let $b_X(B) = b(A, X)$ (idem $\hat b_X$, $b_X^{-1}$, and $\hat b_X^{-1}$).
Assumption \ref{assumption phi} and Corollary \ref{corollary exact flow intervals} imply 
\begin{align}
&{\rm Prob}(Y_{N+1} \in C_{\hat B}|X_{N+1}) \\
& = {\rm Prob}(A_{N+1} \leq \hat b_{X_{N+1}}^{-1}(Q_{\hat B})|X_{N+1}) \\
& = {\rm Prob}(b_{X_{N+1}}^{-1}(B_{N+1}) \leq \hat b_{X_{N+1}}^{-1}(Q_{\hat B})|X_{N+1}) \\
& = {\rm Prob}(B_{N+1} \leq b_{X_{N+1}}\circ \hat b_{X_{N+1}}^{-1}(Q_{\hat B}))
\end{align}
where we drop the conditioning in the last line because, by assumption, $(B_n, X_n) \sim U_B P_X$ for all $X_n$.
The monotonicity of $b_{X_{N+1}}\circ \hat b_{X_{N+1}}^{-1}(B)$ 
%and $Q_{\hat B} = \hat b_{X_{n*}}(A_{n_*})$ for some $n_* \in \{1, \dots, N\}$, we have 
implies $\hat b_{X_{N+1}}^{-1}(Q_{\hat B}) = Q_{\tilde B}$, where $Q_{\tilde B}$ is the sample quantile of $\{\tilde B_n\}$, $\tilde B_n = b_{X_{N+1}}\circ \hat b_{X_{N+1}}^{-1}(\hat B_n)$.
Test and calibration data are not exchangeable because, $B_{N+1}$ and $\tilde B_n$, $n=1, \dots N$, come from different distributions.
The coverage gap can be bounded in terms of the total variation distance between their distributions,  $P_B$ and  $P_{\tilde B}$, i.e.
${\rm d}_{{\rm TV}}(P_{B}, P_{\tilde B}) = \sup_{Z}|P_{B}(Z)- P_{\tilde B}(Z)|$.
Let $(\hat {\rm P}, \tilde B, B')$ define a \emph{maximal coupling} between $\tilde B_1, \dots, \tilde B_N$ and $B_{N+1}$ defined by ${\rm Prob}(\tilde B_n) = \hat {\rm P}(\tilde B)$, $n=1, \dots, N$, and ${\rm Prob}(B_{N+1}) = \hat {\rm P}(B')$.
Then, 
\begin{align}
    &{\rm Prob}(B_{N+1} \leq  Q_{\tilde B}) \\
    & = \hat {\rm P}(B'\leq Q_{\tilde B}, B' = \tilde B) %\nonumber\\& 
    + \hat {\rm P}(B' \leq Q_{\tilde B}, B'\neq \tilde B) \\
    & \geq  \frac{\lceil (N+1)(1 - \alpha)\rceil }{N+1} - \hat {\rm P}(B' \neq \tilde B) ) 
 \end{align}
where the Maximal Coupling Theorem implies (see \cite{lindvall2002lectures, ross2023second} for a proof) 
\begin{align}
\hat {\rm P}(B' \neq \tilde B) = \frac12 {\rm d}_{{\rm TV}}(P_{B_{N+1}}, P_{\tilde B_{n}})
\end{align}
which, in this case, holds for any $n \in \{ 1, \dots, N\}$ because we assume the data objects are i.i.d. 

Assume $\hat b= (1 - \epsilon) b + \epsilon \delta$ and $\hat b^{-1} = (1 - \epsilon) b^{-1} + \epsilon \delta^{-1}$, for all $(A, X)$.
The invertibility of $\hat b$ implies ${\rm Id} = \hat b \circ \hat b^{-1} = (1 - 2\epsilon) {\rm Id} + \epsilon (b \circ\delta^{-1} + b^{-1} \circ\delta ) + \epsilon^2 ({\rm Id} +  \delta \circ\delta^{-1})$, where ${\rm Id}(B) = B$.
Neglecting second-order terms, we have $b \circ\delta^{-1} + \delta \circ b^{-1} = 2 {\rm Id}$, i.e. $\delta^{-1} = 2 b^{-1}  - b^{-1}\circ\delta \circ b^{-1}$ and $\hat b^{-1} = (1 - \epsilon) b^{-1} + \epsilon (2 b^{-1} -b^{-1}\circ\delta \circ b^{-1}) = (1 + \epsilon) b^{-1} - b^{-1}\circ\delta \circ b^{-1}$.
Let $b_X(A) = b(A, X)$ (idem $b^{-1}$, $\hat b$, $\hat b^{-1}$, and $\delta$).
Since $\psi_X(B) = b_{X_{N+1}}  \circ \hat b^{-1}_{X_{N+1}} \circ \hat b_{X} \circ b^{-1}_{X}$ is monotonic, we may interpret it as an NF.
The density of $(\tilde B_n, X_n) \sim P_{\tilde BX}$ is 
\begin{align}
    p(\psi_X(B), X) = \frac{p(B, X)}{|{\rm det} J_{\psi}(B, X)|} =  
    \frac{u(B) p(X)}{|\partial_B \psi_X(B)|} 
\end{align}
where $|\partial_B \psi_X(B)| = \partial_B \psi_X(B)$ because $\psi_X$ is monotonic.   
Then, up to $o(\epsilon^2)$ errors,
\begin{align}
    &{\rm d}_{{\rm TV}}(P_{B}, P_{\tilde B}) \\
    &= \sup_{(B, X)} \left\|u(B)p(B) \left(1 - \frac{1}{\partial_B \psi_X(B)}\right)\right \| \\
    & \leq  \epsilon \sup_{(B, X)} \|u(B)p(X) \| \sup_{(B, X)} \|1 - \partial_B \psi^{-1}_X(B)\| \\
    & = \epsilon \sup_{(B, X)} \|u(B)p(X) \| \\ \nonumber 
    &\quad \times \sup_{(B, X)} \| \partial_A \delta_X \circ \partial_B b^{-1}_X - \partial_A \delta_{X_{N+1}} \circ \partial_B b_{X_{N+1}}^{-1} \| \\
 & \leq  2 \epsilon \sup_{(B, x)} \|u(B)p(X) \| L_\delta L_{b^{-1}} 
\end{align}
where $L_\delta$  and $L_{b^{-1}}$ are the Lipshitz constants of $\delta_X$ and $b^{-1}_X$.
If $U_B = {\rm Uniform}([0, 1])$, $\sup_{(B, X)} \|u(B)p(X) \| = \sup_{X} \| p(X) \|$, which only depends on the marginal density of the covariates over the attribute space.
Hence,    
\begin{align}
    &{\rm Prob}(B_{N+1} \leq Q_{\hat B}) \\
    &\geq \frac{\lceil (N+1)(1 - \alpha)\rceil }{N+1} - \epsilon \sup_{X} \|p(X) \| L_\delta L_{b^{-1}}
\end{align}
$\square$

\end{document}